\documentclass{article}

\usepackage[nonatbib,preprint]{neurips_2020}

\usepackage[utf8]{inputenc} 
\usepackage[T1]{fontenc}    
\usepackage{xcolor}         
\definecolor{custom-blue}{RGB}{6,69,173} 
\usepackage[colorlinks, allcolors=custom-blue]{hyperref}
\usepackage{url}            
\usepackage{booktabs}       
\usepackage{amsfonts}       
\usepackage{nicefrac}       
\usepackage{microtype}      

\usepackage{amsmath,amssymb,mathrsfs} 
\usepackage{algorithm}
\usepackage{algpseudocode}

\usepackage{amsmath,amssymb,mathrsfs} 
\usepackage{mathtools}
\newcommand{\pn}{p_{\mathcal{N}}}
\newcommand{\pd}{p_\mathcal{D}}
\newcommand{\RR}{\mathbb{R}}

\newcommand{\Pro}{\mathcal{P}}
\newcommand{\spt}{\mathrm{spt}}
\newcommand{\ep}{\varepsilon}

\newcommand{\HH}{\mathcal{H}}

\newcommand{\Dep}{D_{\ep}}
\newcommand{\Lexp}{{{\rm L}^{\exp}_{\ep}}}
\newcommand{\dx}{{\rm d}x}
\newcommand{\dy}{{\rm d}y}
\newcommand{\dt}{{\rm d}t}

\newcommand{\pran}[1]{\left(#1\right)}
\newcommand{\brac}[1]{\left\{#1\right\}}

\newcommand{\Rd}{\mathbb{R}^d}
\newcommand{\R}{\mathbb{R}}
\newcommand{\N}{\mathbb{N}}
\newcommand{\sse}{\subseteq}

\newcommand{\vept}{{v^{\ep}_t}}
\newcommand{\rhoept}{\rho^{\ep}_t}
\newcommand{\rhoep}{\rho^{\ep}}
\newcommand{\phiept}{{\varphi^{\ep}_t}}
\newcommand{\psiept}{{\psi^{\ep}_t}}
\newcommand{\psiep}{\psi^{\ep}}
\newcommand{\phiep}{\varphi^{\ep}}

\usepackage{color}
\usepackage{pstricks}
\usepackage{ifthen}
\newrgbcolor{chriscolor}{.7 .1 .1}
\newrgbcolor{augucolor}{.1 .1 .7}
\newrgbcolor{aramcolor}{.1 .7 .1}
\usepackage{csquotes}

\newtheorem{defin}{Definition}

\newtheorem{thm}{Theorem}
\newtheorem{prop}{Proposition}
\newtheorem{exo}{Example}

\usepackage{subfig}

\title{Learning normalizing flows from Entropy-Kantorovich potentials}

\usepackage{authblk}
\makeatletter
\renewcommand\AB@authnote[1]{\rlap{\textsuperscript{\normalfont#1}}}

\makeatother 

\author[*1]{\textbf{Chris Finlay}\thanks{\textsuperscript{*} Equal contribution. Correspondance to
\texttt{christopher.finlay@mcgill.ca}}}
\author[*2]{\textbf{Augusto Gerolin}}
\author[1]{\textbf{Adam M Oberman}}
\author[*1]{\textbf{Aram-Alexandre Pooladian}}
\affil[1]{Department of Mathematics and Statistics, McGill University,
Montr\'eal, Canada}
\affil[2]{Department of Theoretical Chemistry, Vrije Universiteit Amsterdam, Amsterdam, Netherlands}
\renewcommand\footnotemark{}

\begin{document}

\maketitle

\begin{abstract}
We approach the problem of learning continuous normalizing flows from a dual perspective motivated by entropy-regularized optimal transport, in which continuous normalizing flows are cast as gradients of scalar potential functions. This formulation allows us to train a dual objective comprised only of the scalar potential functions, and removes the burden of explicitly computing normalizing flows during training. After training, the normalizing flow is easily recovered from the potential functions.
\end{abstract}

\section{Introduction}
Normalizing flows \cite{Rez15,TabTur13,tabak2010density} are a popular mechanism for
probabilistic modeling and inference, whereby an unknown distribution is
parameterized by a transformation of the standard Normal distribution. Normalizing
flows provide a general framework for defining
flexible probability distributions over continuous random variables, and have been
applied throughout a wide variety of fields, including density estimation (e.g. \cite{GraCeBetSutDuv19,TabTur13}), generative modelling (e.g.
\cite{CheBehDuv19,KigDha18,Oor18}), and variational inference (e.g.
\cite{BerWel18,KinWel16,Rez15,TomWel16}).

Continuous normalizing flows (CNFs) \cite{GraCeBetSutDuv19} construct
normalizing flows through a continuous time-dependent transformation of the data, in which the
transformation is given as the solution operator of a neural ordinary
differential equation (ODE) \cite{CheBehDuv19,Che18,HabRut17, RutHab19}. In this
framework, normalizing flows are parameterized as a flow generated by a
(learned) vector field. Training this vector field can be difficult, and
significant regularization may be necessary to learn well-behaved flows 
\cite{FinJacJorNurObe20,massaroli,LarsSamy20,yan19}.

Here we take a step back and frame CNFs within the lens of Optimal
Transport (OT) theory \cite{peyrebook,vil08}. This is a natural connection,
due to a correspondence between vector fields and the dynamical formulation
of OT \cite{BenBre00}, which we explore below. Indeed, this direct connection was exploited in \cite{FinJacJorNurObe20} and
\cite{LarsSamy20} to speed the training of vector fields for CNFs. 
While this direct approach to linking CNFs with OT theory has yielded promising
improvements to CNF training, the problems of discretizing an ODE and training
the CNF's vector field remain.

In this work,  we will take an indirect route to constructing CNFs, and completely sidestep the need to
solve an ODE generated by a vector field during CNF training. Instead we will use the property that the flows
encountered in OT are \emph{gradients of scalar potential functions}.
Optimization will be done only in terms of these potential functions, without
discretizing an ODE during training. Afterwards, a CNF will
be recovered from the learned potential functions. In a sense, this is an
energy-based modeling perspective \cite{lecun2006tutorial} on CNFs.

To be more explicit, the link between CNFs and OT studied herein relies on the
\textit{entropy-regularized dynamical formulation} of OT, which will allow us to
construct a time dependent curve $\rho_t$ of densities acting as the
displacement of $\mu$, the data measure, to
$\nu$, the Gaussian measure. More precisely, the entropy-regularized dynamical OT problem seeks the  pair $(\rho_t, v_t)$ of density-flow $\rho_t$ and vector field $v_t$ minimizing the variational problem
\begin{align}\label{eq: intro_p}
  \underset{(\rho_t, v_t)}{\inf}  \int_0^1 \int_{\Rd} \pran{\frac{\|v_t\|^2}{2} +
  \frac{\ep^2}{8}\| \nabla \log \rho_t \|^2}\rho_t \dx \dt,
\end{align}
subject to the constraint that the pair also satisfies the \textit{continuity equation} 
\begin{align}\label{eq:intro_ce}
\partial_t\rho_t + \nabla\cdot(v_t\rho_t) = 0,
\end{align}
and that both $\rho_0=\mu$ and $\rho_1=\nu$ (i.e. that the initial and final
endpoints respectively agree). The scalar $\ep$ will control the amount of regularization provided by the
Fisher information term $\| \nabla \log \rho_t\|^2$, and corresponds to entropic
regularization of OT \cite{Cut13}. 

Entropic regularization will play an important role in our approach. In
particular,
since in general we do not have access to the true data distribution, this type
of regularization introduces inherent stochasticity to the learned flows,
which may heuristically be beneficial during training and inference. 
In addition, entropic regularization will simplify our numerical method by
allowing us to approximate a particular function transformation with Monte-Carlo integration.
Notice that when $\ep=0$, the variational problem \eqref{eq: intro_p} selects,
among all pairs $(\rho_t,v_t)$,  the one that minimizes the kinetic energy of the vector
field. 
\begin{figure}[t]
\captionsetup[subfigure]{labelformat=empty}
\centering
\subfloat[$t=0$]{%
\includegraphics[width=0.23\textwidth]{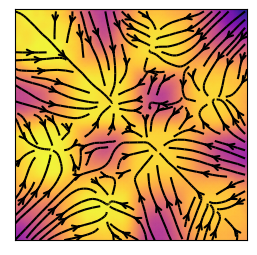}%
}\hfil
\subfloat[$t=\frac{1}{3}$]{%
\includegraphics[width=0.23\textwidth]{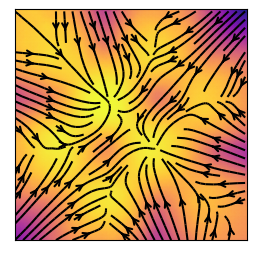}%
}\hfil
\subfloat[$t=\frac{2}{3}$]{%
\includegraphics[width=0.23\textwidth]{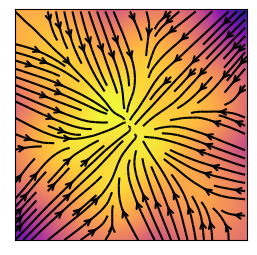}%
}\hfil
\subfloat[$t=1$]{%
\includegraphics[width=0.23\textwidth]{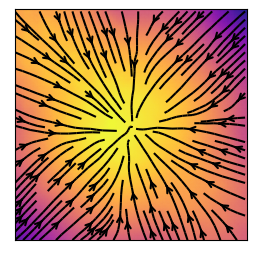}%
}
\label{fig: intro}
\caption{Using the learned Entropy-Kantorovich potentials and \eqref{eq:
punchline}, the vector field $v_t$ (black arrows) recovered from the potentials
creates a CNF between the
checkerboard distribution (at $t=0$) and the standard Normal distribution (at
$t=1$). Log-densities of the
distributions along the flow are shown with the heat map.}
\end{figure}

It is from the continuity equation \eqref{eq:intro_ce} that a CNF is defined,
by the ODE 
\begin{align}\label{eq:intro_ode}
 X'_t = v_t(X_t(x)) \quad \text{s.t.} \quad X_0(x) = x; \quad x \in \R^d
\end{align}
where $x$ is a particle drawn from the data.
The continuity equation can be interpreted as the equation ruling the evolution of 
a family of particles initially drawn from the data measure $\mu$, and
where each particle follows the path defined by the solution operator of
\eqref{eq:intro_ode}, flowing the particles to the Normal distribution.

\textbf{Main contributions} 

In practice, directly optimizing the variational problem \eqref{eq: intro_p} may
not be feasible in high dimensions, and optimizing the CNF generated by
\eqref{eq:intro_ode} introduces its own difficulties. In this paper we instead
use theoretical results from OT theory
to provide implicit formulas for the optimal flow $(\rho_t,v_t)$ solving
\eqref{eq: intro_p}. The flow so defined will only depend on two scalar
functions $\varphi$ and $\psi$, called \textit{Entropy-Kantorovich potentials}.
The CNF will be defined through the following vector field
\begin{align}\label{eq: punchline}
  v_t(x) = \nabla\frac{1}{2} (\varphi_t(x) - \psi_t(x)), 
\end{align} 
where $\varphi_t:= \ep \log\HH_{t\ep} [e^{\varphi/\ep}]$ and
$\psi_t:=\ep \log\HH_{(1-t)\ep} [e^{\psi/\ep}]$, with $\HH$ the
heat kernel (see Section \ref{sec: bridge}; $\HH$ is also known as Gaussian
averaging). Additionally, the density-flow
will be defined by 
\begin{align}
  \log\rho_t(x) = (\varphi_t(x) + \psi_t(x))/\varepsilon.
\end{align}
We will show how the Entropy-Kantorovich potentials can be found by optimizing a static
\textit{dual problem} of \eqref{eq: intro_p}; the optimization procedure itself
will not require the numerical solution of an ODE. 

To summarize, our main contributions are the following:
\begin{itemize}
  \item We introduce a novel framework for constructing CNFs from potential
    functions, based on ideas from entropy-regularized OT. The framework is
    theoretically well motivated, and interprets the CNF so defined as a curve
    in the 2-Wasserstein space of probability measures, connecting the data to
    the Normal distribution.
  \item Our method is computationally efficient: training is sample-based and
    mesh-free, and only requires optimizing two time-invariant
    Entropy-Kantorovich potential
    functions $\varphi$ and $\psi$. The method completely avoids solving an ODE
    during training.
  \item Once the potential functions have been trained, it is straightforward to
    perform density estimation and generative modeling through a CNF recovered
    from the learned potential functions. The CNF is
    evaluated easily, and can be applied in higher dimensions through the use of
    Monte-Carlo integration.
  \end{itemize}

\subsection{Related work} 
The tools provided by the literature on OT
\cite{peyrebook,vil08} are the cornerstone of our methodology. Connections
between particle-based methods in numerical analysis and OT first
appeared in the seminal work of \cite{BenBre00}, and is referred to as the
``Benamou-Brenier'' formulation in the modern literature. This provides a \textit{dynamic} perspective on OT, which can be generalized to entropic optimal transport (EOT) \cite{gentil2017, GigTam18, LeoSurvey}. These connections will be made more explicit in Section {\ref{sec:CFNEOT}}.

Meanwhile, the literature on flow-based methods in deep learning is rich with
applications of OT theory, particularly in the context of normalizing flows;
examples include \cite{PavTabTri18,RutHab19,Ruthotto9183, TriTab16,XanWeiWan18}. A relevant
connection to our work appears in \cite{FinJacJorNurObe20}, where the authors
exploit the Benamou-Brenier formulation by adding the relevant kinetic energy
term to the objective function. Following work in \cite{LarsSamy20} cast the
vector field as a gradient potential. However, in both these works the authors
directly solve the ODE generated by
the time-dependent vector field during training, which is in contrast to our approach.

In the context of Wasserstein Generative Adversarial Networks (GANs), different
approaches have been taken to learn Kantorovich potentials using neural networks; see for example \cite{arjovsky17, gulrajani17}, and the
entropy-regularized case in \cite{mallasto19}.

\section{CNFs from entropic optimal transport}\label{sec:CFNEOT}
\textbf{Notation:}
$\Pro(\Rd)$ denotes the set of probability distributions over $\R^d$, with
$\Pro_p(\R^d)$ being the set of probabilities with $p\geq 1$ finite moments. The density of the standard Normal distribution in $\Rd$ is denoted 
$\pn$ and the data measure is denoted by $\pd$. In the OT framework below, we
identify $\pn$ with the target measure $\nu$, while $\pd$ is the source
measure $\mu$. In principle, we could let the target measure be any closed-form
density function, but in this work we
always take $\nu$ to be the standard Gaussian measure, as done in the
normalizing flow literature.  We sometimes refer to $\pn$ ($\nu$) as the Normal distribution, despite it being a measure.

The $2$-Wasserstein distance between $\mu,\nu \in \Pro_2(\Rd)$ is defined as 
\begin{equation}\label{eq:W2}
W_2(\mu,\nu) := \pran{\underset{\gamma \in \Pi(\mu,\nu)}{\min} \int_{\R^d\times\R^d} \frac{1}{2}\| x-y \|^2_2 \ d\gamma(x,y)}^{1/2},
\end{equation}
where $\Pi(\mu,\nu)$ denotes the set of probability measures on $\Rd \times
\Rd$ with marginals equal to $\mu$ and $\nu$,
$$\Pi(\mu,\nu) := \brac{\gamma \in \Pro_2(\Rd\times\Rd) \ \big| \ \gamma(A \times \R^d) = \mu(A), \gamma(\R^d \times A) = \nu(A) }.$$
An element in $\Pi(\mu,\nu)$ is called a coupling or a transport plan, and
$\gamma^{\text{opt}}\in\Pi(\mu,\nu)$ realising the minimum in \eqref{eq:W2} is
called the optimal transport plan. For a map $T:\R^d\to\Rd$,
$T_{\sharp}\mu$ denotes the \textit{pushforward} of $\mu$ with $T$, i.e.
$\mu(T^{-1}(A)) = T_{\sharp}\mu(A)$, for any $A \sse \R^d$ Borel
measurable\footnote{We refer the reader to Appendix \ref{app:OT} for background
on optimal transport maps.}.

The space $\Pro_2(\Rd)$ endowed with $W_2$ is a complete and separable metric space, denoted by $\mathbb{W}_2(\R^d)$. We can also show that $\mathbb{W}_2(\R^d)$ is a geodesic space, i.e. any two points in $\mathbb{W}_2(\R^d)$ can be connected by a geodesic\footnote{A geodesic is a curve that minimizes the ``length" between two end-points.} (see Appendix \ref{app:geodesics} for details).

Finally, we define the function space $\Lexp(\Rd;\dx) := \{ f :\Rd \to
\R \ | \  \int_{\Rd}\exp(f/\varepsilon) \dx < +\infty \}$ and the $(c,\varepsilon)$-transformation of $f \in \Lexp(\Rd;\dx)$: 
\begin{equation}\label{eq:cep}
  f^{(c,\ep)}(y;\nu) := \ep\log\left({\frac{1}{(2\pi
  \varepsilon)^{\frac{d}{2}}}}\int_{\Rd}e^{\frac{f(x)-\frac{1}{2}\| x-y\|^2}{\ep}}\dx \right) - \ep\log(\nu(y)), 
\end{equation}
and analogously for $g \in \Lexp(\Rd;\dy)$ (with $\nu$ swapped for $\mu$). This
transformation is critical to our method. It arises throughout many branches of
math and physics under various names: (i) the Hopf-Cole transformation of $f$, and is the solution
operator of a partial differential equation from stochastic control theory; (ii)
the softmax operation of $f$ convolved with a Normal distribution of variance
$\varepsilon$; (iii) additionally the $(c,\varepsilon)$-transform is a smoothed version of the quadratic $c$-transform arising in OT theory and convex analysis.
The $(c,\ep)$-transform lends itself to evaluation in high dimensions or in mesh-free environments by Monte-Carlo integration, whereas the $c$-transform is difficult to compute in these scenarios.
Both the $(c,\varepsilon)$-transform and the space $\Lexp(\Rd;\dx)$ are intimately tied to energy-based models (see Equation
\eqref{eq:SchSys} below).

\subsection{Transport maps and normalizing flows}\label{sec: nfs} 
Here we recall some basic facts about normalizing flows and transport maps, with a clearer exposition in Appendix \ref{app:OT}.
Under some regularity assumptions on $\mu,\nu$, the optimal transport map $T$ is
$1$-Lipschitz \cite{Caf00,Caf02} and then, $T_{\sharp}\mu=\nu$ can be
equivalently written by the change of variables formula $\mu(x) = \nu(T(x))\vert
\mathrm{det}\,\mathrm{J}_T(x)\vert$, where $\mathrm{J}_T$ is the Jacobian of $T$. Therefore, when
$\mu=\pd$ is the data density and $\nu=\pn$ is the standard Normal density, $T$ is a normalizing flow.

Normalizing flows are designed to maximize the log-likelihood of the data under
a  transformation $T$. The normalizing flow so defined is not necessarily an
optimal transport map. Since the data density is not known analytically, the
difficulty of evaluating the log-likelihood of a sample $x$ under $\pd$ is
pushed onto evaluating the log-likelihood of $T(x)$ under $\pn$:
\begin{equation}\label{eq:log-lik}
  \log \pd(x) = \log \pn(T(x)) + \log \vert \mathrm{det}\,\mathrm{J}_T(x)\vert.
\end{equation}

In the
normalizing flow literature, $T$ is a composition of `simple' analytic
functions, so that the log-determinant of the Jacobian can be computed
tractably. For example, in CNFs, where the transport map is defined as the
solution operator of the ODE \eqref{eq:intro_ode}, the Jacobian log-determinant
is evaluated by integrating the divergence of the vector field $v_t$ along the
solution path \cite{GraCeBetSutDuv19}. Once a family of maps with tractable
Jacobian determinants have been constructed, and given data $x_i \sim \pd$, the objective of normalizing flows is to
simply maximize the log-likelihood of the data,  $\max_T
\sum_i \log
\pn(T(x_i)) + \log \vert \mathrm{det}\,\mathrm{J}_T(x_i)\vert$.

\subsection{Geodesics flows in the $2$-Wasserstein space $\mathbb{W}_2(\Rd)$}\label{sec: geo1}
In this section, we briefly discuss how to construct a CNF which is a
\textit{constant-speed geodesic} between $\pd$ and $\pn$ in $\mathbb{W}_2(\Rd)$,
deferring technical details to Appendix \ref{app:geodesics}.

Suppose there exists an optimal
transport map $T$ between $\pd$ and $\pn$, and consider the convex combination
between the identity map $\text{Id}$ and $T$, $\pi_t(x) = (1-t)x + tT(x)$. This can be viewed as a interpolant between our two measures of interest. In fact, the continuous deformation $\rho_t = (\pi_t)_{\sharp}\pd$, for $t \in
[0,1]$, is a constant-speed geodesic between the data and the
Normal distribution.

By the celebrated Brenier's Theorem \cite{Bre91}, the optimal transport map $T$ can be expressed as the gradient of a scalar-valued potential $\phi:\R^d\to\R$, $T(x) = \nabla (\frac{1}{2}\|x\|^2 - \phi(x))$, then $\rho_t = (\pi_t)_{\sharp}\pd$ reads 
\begin{equation}\label{eq: rhot}
\rho_t = ({\rm Id} + t\nabla\phi(x))_{\sharp}\pd.
\end{equation}
In ODE terms, the velocity field defining the ODE \eqref{eq:intro_ode} of this continuous normalizing flow is given by
$v_t(x) = T(x)-x = \nabla\phi(x)$, and is \textit{time-invariant}, depending only on the point $x$,
and is hence constant-speed. 
The function $\phi$ is called a \textit{Kantorovich potential} and is related to the dual problem of \eqref{eq:W2}. 
One can verify that $\rho_t$ solves the continuity equation $\partial_t \rho_t +
\nabla\cdot(\nabla\phi \rho_t) = 0$ from which the continuous normalizing flow
is read off. In the context of normalizing flows, the continuity equation
dictates
the evolution of data moving from  $\pd$ to $\pn$, if the
paths were to truly take the optimal trajectory. 

\subsection{Entropy-regularized $2$-Wasserstein distance}
While the approach of Section \ref{sec: geo1} is elegant, it is difficult to
optimize the Kantorovich potential directly $\phi$ in a mesh-free environment, or
in high-dimensions. We instead turn to the entropy-regularized optimal
transport problem, which as we shall see, lends itself to a computationally
tractable method to determine the potential functions.

The entropic regularization of the $W_2$ distance with regularization parameter
$\varepsilon > 0$ \cite{Cut13} is defined by  
\begin{equation}\label{eq:Wep}
W^2_{\ep}(\pd,\pn) =  \min_{\gamma\in\Pi(\pd,\pn)}\int_{\Rd\times \Rd}\frac{1}{2}\| x-y\|^2 d\gamma(x,y) + \ep {\rm H}(\gamma).
\end{equation}
Here ${\rm H}$ is the entropy of a probability measure $\gamma \in \Pro_2(\R^d\times\R^d)$, defined by ${\rm H}(\gamma) =
\int_{\R^d\times\R^d} \gamma(x)\log\pran{\gamma(x,y)}\dx\dy$ if  $\gamma$ is a density and
${\rm H}(\mu) = +\infty$ otherwise. By strong convexity \eqref{eq:Wep} always
admits a \textit{unique} minimizer \cite{peyrebook, FraLor89, LeoSurvey}.
Entropic regularization has the effect of `diffusing' or `fuzzing' the transport
plans.

An equivalent formulation of \eqref{eq:Wep}, the so-called dual or Entropy-Kantorovich formulation of \eqref{eq:Wep} allows us to obtain the distance between $\pd$ and $\pn$ by maximizing over
pairs of Entropy-Kantorovich potentials $(\psi, \varphi)$ rather than minimizing over measures $\gamma$
\cite{peyrebook,DMaGer19,LeoSurvey},
\begin{equation}\label{eq:dualep}
  W_{\ep}^2(\pd,\pn) = \sup\left\{ \Dep(\varphi,\psi) : \varphi \in \Lexp(\Rd;\dx), \psi\in\Lexp(\Rd;\dy) \right\} + \ep,
\end{equation}
where $\Dep:\Lexp(\Rd;\dx)\otimes\Lexp(\Rd;\dy)\to \RR$ is the dual functional 
\begin{align}\label{eq:dep}
\begin{split}
  \Dep(\varphi,\psi) = \int_{\Rd}\varphi(x) &\,\mathrm{d}\pd(x) + \int_{\Rd}\psi(y)
\,\mathrm{d}\pn(y)  \\ 
&- \ep\int_{\Rd\times\Rd}\exp\left\{\frac{\varphi(x)+\psi(y)-\frac{1}{2}\| x-y\|^2}{\ep} \right\}\dx\dy.
\end{split}
\end{align}
Note that the functional $\Dep$ is strictly concave in each variable and, under
mild hypotheses, one can show the existence of maximizers in \eqref{eq:dualep}
which are unique up to additive constants \cite{DMaGer19}. Useful
characterizations of the primal \eqref{eq:Wep} and dual problem \eqref{eq:dep}
are given by the following theorem.

\begin{thm}[Proposition 2.11, \cite{DMaGer19}]\label{teo:entropic}
Let $\ep>0$ be a positive number, $\Omega\subset\Rd$ be a compact set, $\pd,\pn \in \Pro(\Omega)$. Then given $\varphi \in \Lexp(\Rd;\dx)$ and $\psi \in \Lexp(\Rd;\dy)$, the following are equivalent:
\begin{enumerate}
\item \textit{(Maximizers)} $\varphi$ and $\psi$ are maximizing potentials for \eqref{eq:dualep};
\item \textit{(Maximality condition)} $\varphi^{(c,\ep)}=\psi$, $\psi^{(c,\ep)}=\varphi$ and, moreover, $ \varphi,\psi\in L^{\infty}(\Omega)$.
\item \textit{(Primal problem)}  $\gamma^{\text{opt}}_{\ep} = \exp\left((\varphi(x)+\psi(y)-\frac{1}{2}\| x-y\|^2)/\ep\right) \in \Pi(\pd, \pn)$; 
\item \textit{(Duality attainment) }$W^2_{\ep}(\pd,\pn) = \Dep(\varphi,\psi) +\ep$.
\end{enumerate}
Moreover, the optimal coupling $\gamma^{\text{opt}}_{\ep}$ is also the (unique) minimizer for the problem \eqref{eq:Wep}.
\end{thm}

When $\psi$ and $\varphi$ are optimal, we may read off the data and normal
log-densities from the Entropy-Kantorovich potential functions:
\begin{align}\label{eq:SchSys}
\begin{dcases}
  \varphi(x) + \psi^{(c,\varepsilon)}(x) =  \varepsilon \log\pd(x) + C_1\\
  \psi(y) + \varphi^{(c,\varepsilon)}(y) = \varepsilon \log \pn(y) + C_2
\end{dcases}
\end{align}
for some normalizing constants $C_1$ and $C_2$. In other words, the potential functions
parameterize the data and Normal distributions as energy-based models.

\subsection{A bridge between CNFs and potentials: the dynamic
formulation}\label{sec: bridge} 
We are now in a position to bridge entropic optimal transport with continuous
normalizing flows. The
variational problem \eqref{eq:Wep} can be expressed in a \textit{dynamic}
form {\rm \cite{GigTamBB18,LeoSurvey}
\begin{align}\label{eq: fisher_dual}
\frac{\ep}{2}\left({\rm H}(\pd) + {\rm H}(\pn)\right) + \underset{(\rho_t^\ep, \vept)}{\inf} \int_0^1\int_{\Rd}   \pran{\frac{\|\vept\|^2}{2} + \frac{\ep^2}{8}\| \nabla \log \rho_t^\ep \|^2}\rho_t^{\ep} \dx\dt ,
\end{align}
with the constraint that $(\rho_t^\ep, \vept)$ solves the continuity equation
$\partial_t \rho_t^\ep + \nabla\cdot(v^{\ep}_t\rho^{\ep}_t)=0$, and that
$\rhoep_{0} = \pd$, $\rhoep_1 = \pn$}. The time-dependent density $\rhoep_t$ is a
curve between the data and Normal distributions in the $2$-Wasserstein space
parameterized by $t\in[0,1]$.
Once $v^\ep_t$ is known, this time dependent vector field defines a CNF
via the ODE \eqref{eq:intro_ode}.

Equation \eqref{eq: fisher_dual} also has an associated dual problem (equivalent
  to \eqref{eq:dep}), where again instead of
minimizing over pairs $(\rhoep_t, v^\ep_t)$,  optimization takes place across the
following two functionals:
\begin{align}\label{eq: dyn_dual}
\begin{dcases}
  &J(\varphi) = \varepsilon H(\pd) + \sup_\varphi \int_{\Rd} \varphi \,
  \mathrm{d}\pn
\,+ \int_{\Rd} \varphi^{(c,\ep)} \,\mathrm{d}\pd, \\
&I(\psi) = \varepsilon H(\pn) +  \sup_\psi \int_{\Rd} \psi \,\mathrm{d}\pd \,
+ \int_{\Rd} \psi^{(c,\ep)} \,\mathrm{d}\pn,
\end{dcases}
\end{align}
$\varphi^{(c,\ep)}$ and $\psi^{(c,\ep)}$ are, respectively the
$(c,\ep)$-transforms of $\varphi$ and $\psi$ defined in \eqref{eq:cep}. 
We refer the reader to \cite{GigTam18} for a derivation of this result.
In practice, it is through \eqref{eq: dyn_dual} that we will build our numerical
method: we will solve for $\varphi$ and $\psi$, after which the CNF will be
recovered.

\paragraph{Recovering the flow and density:} We first define the convolution operator
\begin{equation}\label{eq: h_smoothing}
\HH_s[f](y) := \frac{1}{(2\pi s)^\frac{d}{2}} \int_{\Rd} f(x) \exp\left(-\frac{1}{2s}\|x-y\|^2\right) \dx,
\end{equation}
which smooths the operand with the Normal distribution of variance $s$; this is
sometimes called the heat kernel. Note the similarities with the
$(c,\ep)$-transform. Let $(\varphi,\psi)$ be the optimal
Entropy-Kantorovich potentials in \eqref{eq: dyn_dual}, and define $\varphi_t:=
\ep \log\HH_{t\ep} [e^{\varphi/\ep}]$ and
$\psi_t:=\ep \log\HH_{(1-t)\ep} [e^{\psi/\ep}].$ Then the \textit{entropic-displacement interpolation} $\rhoep_t:[0,1]\to\Pro_2(\Rd)$ between the probability densities $\pd$ and $\pn$ and the corresponding velocity field $v^{\ep}_t$ are given by \cite{LeoSurvey}
\begin{equation}\label{eq:rhot_ep}
  \rho_t^\ep(x) = \exp\pran{(\varphi_t(x) + \psi_t(x))/\ep}, \mbox { and }
\end{equation}
\begin{equation}\label{eq: vt_ep}
 v^\ep_t(x) = \nabla \left(\varphi_t(x) - \psi_t(x)\right)/2,
\end{equation}
The entropic interpolant $\rho^\ep_t$ given by \eqref{eq:rhot_ep} is the
regularized analogue to the constant speed geodesic defined in Section
\ref{sec: geo1}. Moreover, as $\ep\to 0$, $\rhoep_t\to \rho_t$, the
$2$-Wasserstein geodesic between $\pd$ and $\pn$ \eqref{eq: rhot} introduced in
Section \ref{sec: geo1} (see e.g. \cite{LeoSurvey}). In Appendix
\ref{app:dynOTep}, we illustrate the smoothing effect of entropic regularization
on the $2$-Wasserstein geodesic between two Gaussian distributions, where a closed-form solution is known.

We emphasize that once $\varphi$ and $\psi$ are known, we can completely defined a
continuous normalizing flow between the data distribution and the standard
Normal via equations \eqref{eq: vt_ep} and the ODE \eqref{eq:intro_ode}.

\section{Numerics}\label{sec:num}
We now have the necessary tools to build CNFs by solving for Entropy-Kantorovich
potentials. We parameterize the pair of Entropy-Kantorovich potentials
$(\varphi,\psi)$ as neural
networks  $(\varphi_\theta,\psi_\omega)$ with respective parameters $\theta$ and
$\omega$. We solve the dual problem \eqref{eq:dualep} by maximizing the pair of
functionals \eqref{eq: dyn_dual} over batches sampled from $\pd$ and $\pn$.
The complete
pseudo-code of our training procedure is outlined in Algorithm \ref{alg:main}.

\paragraph{Alternating between optimizing $\varphi$ and $\psi$}
In practice, we take alternating gradient ascent steps on the functional $J$ in
$\varphi$ and the functional $I$ in $\psi$. This alternating approach is motivated
by the following.
\begin{prop}[Lemma 2.6 in \cite{DMaGer19}]\label{lemma:asccep} The dual function $\Dep: \Lexp(\Rd;\dx) \times \Lexp(\Rd;\dy)\to\mathbb{R}$ defined as in \eqref{eq:dep} is concave in each one of the variables. Moreover,
\begin{itemize}
\item[] $D_{\ep}(\varphi,\varphi^{(c,\ep)}) \geq  D_{\ep}(\varphi,\psi), \,
  \forall \, \varphi \in \Lexp(\Rd;\dx)$, 
\item[] $D_{\ep}(\varphi,\varphi^{(c,\ep)}) =  D_{\ep}(\varphi,\psi) \text{ if and only if } \psi = \varphi^{(c,\ep)}.$
\end{itemize} 
In particular we can say that $\varphi^{(c,\ep)} \in {\rm argmax} \{ D_{\ep} (\varphi,\psi) \; : \; \psi \in \Lexp(\Rd;\dx) \}$. Clearly, an analogous results holds by exchanging the roles of $\varphi$ and $\psi$.
\end{prop}
In other words, we can create an increasing sequence of objective values by alternating between placing only $\varphi$ and only $\psi$ (with
their respective $(c,\ep)$-transforms in place of the other potential function) in the arguments of
$D_\ep$. We note that this sequence of objective  
function values is increasing only up to error induced by mini-batch sampling.
An analysis of this error is outside the scope of this paper.

Then, at optimum, Theorem
\ref{teo:entropic} tells us that the optimal potentials in \eqref{eq:dualep}
satisfy $\varphi^{(c,\ep)} = \psi$ and $\psi^{(c,\ep)} = \varphi$. 
Moreover, $D_\ep(\varphi,\psi)$ is bounded above by the
entropy-regularized functional $W_{\ep}(\pd,\pn)$, since $\Dep(\varphi,\psi) \leq W_{\ep}(\pd,\pn) - \ep, \, \forall \, \varphi,\psi$ (see also Lemma 2.10 in \cite{DMaGer19}).

\paragraph{Fast approximate $(c,\ep)$-transform} The $(c,\ep)$-transformation building the
above sequences can be approximated efficiently via Monte-Carlo (MC) integration with $N$
samples  $x_i \sim \mathcal N(y,\ep)$:
\begin{align}\label{eq:estimation}
(\varphi_{\theta})^{(c,\ep)}(y) &=
\ep\log\left(\frac{1}{(2\pi\ep)^{\frac{d}{2}}}\int_{\Rd}e^{\frac{\varphi_{\theta}(x)-\frac{1}{2}\|
x-y \|^2}{\ep}}\dx\right) \approx \ep\log\left(\frac{1}{N}\sum^N_{i=1}e^{\varphi_{\theta}(x_i)/\ep}\right).
\end{align}
Monte-Carlo integration is well known to be close the true integral
point-wise with an error of $\mathcal{O}(N^{-1/2})$ in the number of samples (for fixed dimension $d$) \cite{robert2013monte}. We can safely omit the second term in \eqref{eq:cep}, as we are only interested in the argmax of the objective function, and not the optimal function value.
We will also use MC integration for a fast evaluation of the heat kernel $\HH$.

\paragraph{Constructing the CNF and the velocity field $v_t$}
Finally, upon optimizing for $\varphi_\theta$ and $\psi_\omega$, the optimal
vector field generating the CNF is given by \eqref{eq: vt_ep}.
The CNF framework \cite{GraCeBetSutDuv19} allows us to both estimate probability density
and generate samples. For a given $x_i\in\mathcal D$, the log-likelihood of the
data point is
computed via \eqref{eq:log-lik}, where the transformation is provided by solving
\eqref{eq:intro_ode}. Generation is done by sampling $z_i \sim \mathcal N(0,1)$
and running \eqref{eq:intro_ode} backwards in time.
Note that because we use MC integration for the heat kernel, computation of
$v_t$ is mesh-free, quick, and scales easily to high dimensions. 
\begin{algorithm}[b]
  \caption{Dual ascent of potential functions, parameterized by neural networks}
  \label{alg:main}
  \begin{algorithmic}
    \State Input: Target dataset $\mathcal{D} \sse \Rd$, $\varepsilon > 0$;
    $N,\,B,\,k_{\max} \in \N$; $k = 0$ and step-size $\eta > 0$
	  \State Initialize networks $\varphi_{(0)}, \psi_{(0)}$ 
		\While{$k < k_{\max}$} 
		\For{$x_B \in \mathcal{D}$}
    \State Sample $z_B \sim \mathcal{N}(0,I_d)$  \Comment{Sampling from $\pn$}
		\State Compute $\varphi^{(c,\ep)}_{(k)}$ and $\psi^{(c,\ep)}_{(k)}$ with
    MC integration, using $N$ samples
		\State Update $\varphi_{(k)}$ using a stochastic optimizer over data $(x_B,z_B)$, with $\psi_{(k)}$ fixed:
         $$\varphi_{(k+1)} \leftarrow \varphi_{(k)} + \eta\nabla \tilde{J}(\varphi_{(k)})$$ 
        \State Update $\psi_{(k)}$ using a stochastic optimizer over data $(x_B,z_B)$, with $\varphi_{(k+1)}$ fixed:
         $$\psi_{(k+1)} \leftarrow \psi_{(k)} + \eta\nabla \tilde{I}(\psi_{(k)})$$
		\EndFor
		\State $k \leftarrow k+1$
        \EndWhile
  \end{algorithmic}
\end{algorithm}

\paragraph{Speeding optimization by reinforcing the $(c,\ep)$-transform}
In practice we have found optimization is helped by reinforcing the constraint
that $\varphi^{(c,\ep)} = \psi$ and $\psi^{(c,\ep)} = \varphi$.
To do so, we re-define the
objective dual function \eqref{eq:dep} with an extra auxiliary variable
\begin{align}
\Dep(\varphi,\tilde{\varphi},\psi) = \int_{\Rd}\varphi d&\pd +
\int_{\Rd}\tilde{\varphi} d\pn
-\ep\int_{\Rd\times\Rd}\exp\left\{\frac{\varphi+\psi-\frac{1}{2}\| x-y\|^2}{\ep}
\right\}\dx\dy \nonumber.
\end{align}
Optimization then alternates over the twin functionals $\tilde{J}$ and $\tilde{I}$,
which are motivated by
Proposition \ref{lemma:asccep} and equation \eqref{eq: dyn_dual}
\begin{align}
& \tilde{J}(\varphi_\theta) := \Dep(\varphi_\theta,
\varphi_\theta^{(c,\ep)},\psi_\omega) + \alpha\Vert
\varphi_\theta^{(c,\ep)}-\psi_\omega \Vert^2, \\
& \tilde{I}(\psi_\omega) := \Dep(\psi_\omega, \psi_\omega^{(c,\ep)},
\varphi_\theta) + \alpha\Vert \psi_\omega^{(c,\ep)}-\varphi_\theta \Vert^2.
\end{align}
We have incorporated
an additional $L_2$-regularization term with strength $\alpha>0$
for extra reinforcement of the optimality conditions over mini-batches.

\paragraph{Examples}
We consider several low-dimensional distributions commonly used in the
normalizing flow literature, some of which are discontinuous (e.g.
checkerboard). For these experiments, we parameterize the two
Entropy-Kantorovich potential functions $(\varphi_\theta,\psi_\omega)$ using
four fully connected linear layers with ReLU activations, with hidden dimension
64. The hyper-parameters for the experiments are provided in Appendix \ref{sec:
algo_main} and, apart from the total number of iterations, are the same for each
dataset. Indeed, we observed that some of the distributions were more difficult
to model than others, and needed more time to optimize over the function space.

\begin{figure}[b]
\captionsetup[subfigure]{labelformat=empty}
\centering
\subfloat{%
  \rotatebox{90}{\parbox{2cm}{\centering Ground truth\\$\log \pd$}}
\includegraphics[width=0.12\textwidth]{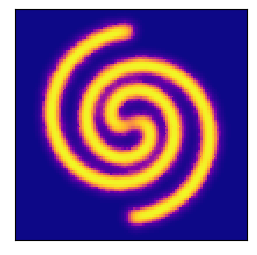}%
}
\subfloat{%
\includegraphics[width=0.12\textwidth]{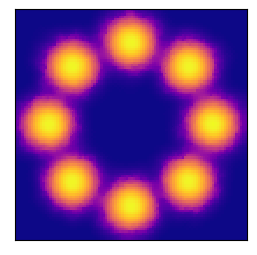}%
}
\subfloat{%
\includegraphics[width=0.12\textwidth]{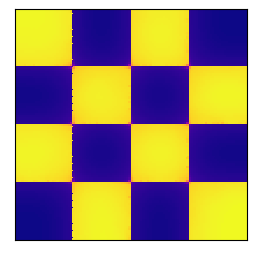}%
}
\subfloat{%
\includegraphics[width=0.12\textwidth]{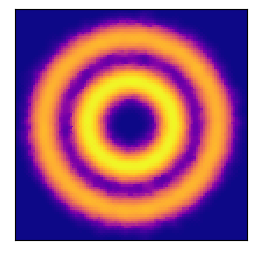}%
}
\subfloat{%
\includegraphics[width=0.12\textwidth]{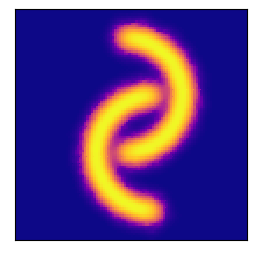}%
}
\subfloat{%
\includegraphics[width=0.12\textwidth]{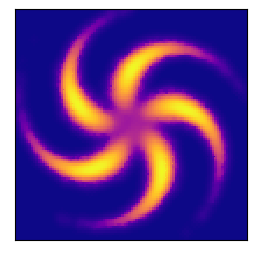}%
}
\subfloat{%
\includegraphics[width=0.12\textwidth]{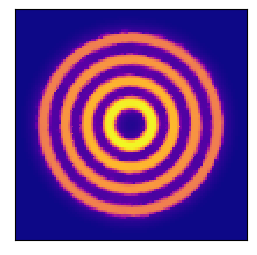}%
}
\subfloat{%
\includegraphics[width=0.12\textwidth]{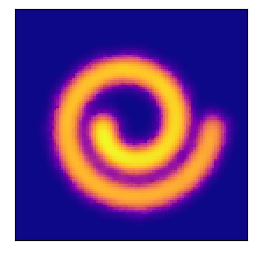}%
}\vspace{-4mm}\\
\subfloat{%
  \rotatebox{90}{\parbox{2cm}{\centering Estimated\\$\log \rho_0^\ep$}}
\includegraphics[width=0.12\textwidth]{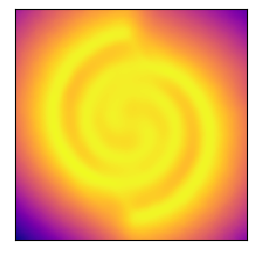}%
}
\subfloat{%
\includegraphics[width=0.12\textwidth]{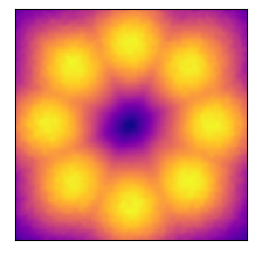}%
}
\subfloat{%
\includegraphics[width=0.12\textwidth]{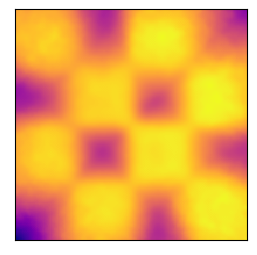}%
}
\subfloat{%
\includegraphics[width=0.12\textwidth]{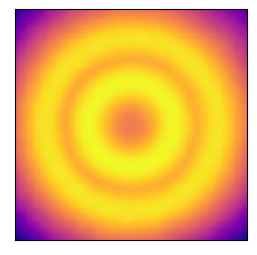}%
}
\subfloat{%
\includegraphics[width=0.12\textwidth]{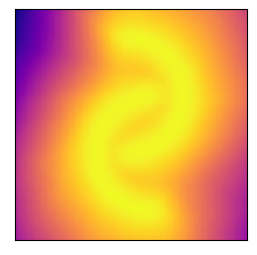}%
}
\subfloat{%
\includegraphics[width=0.12\textwidth]{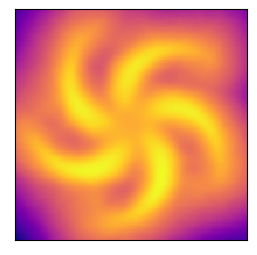}%
}
\subfloat{%
\includegraphics[width=0.12\textwidth]{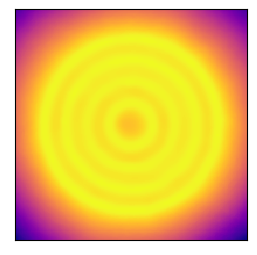}%
}
\subfloat{%
\includegraphics[width=0.12\textwidth]{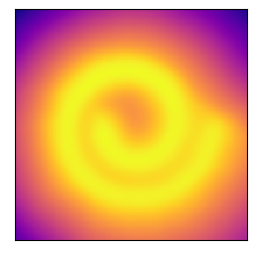}%
}\vspace{-4mm}\\
\subfloat{%

  \rotatebox{90}{\parbox{2cm}{\centering Samples\\$x\sim\rho_0^\ep$}}
\includegraphics[width=0.12\textwidth]{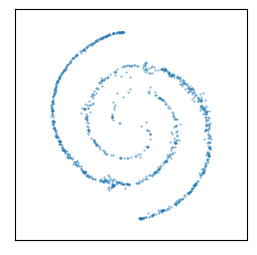}%
}
\subfloat{%
\includegraphics[width=0.12\textwidth]{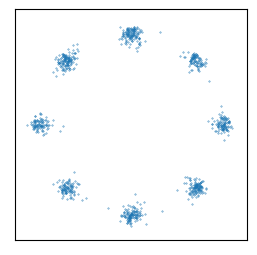}%
}
\subfloat{%
\includegraphics[width=0.12\textwidth]{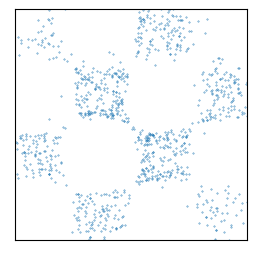}%
}
\subfloat{%
\includegraphics[width=0.12\textwidth]{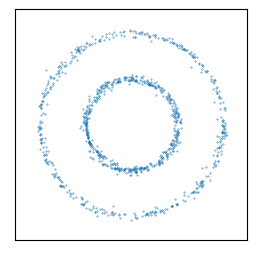}%
}
\subfloat{%
\includegraphics[width=0.12\textwidth]{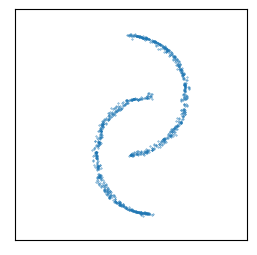}%
}
\subfloat{%
\includegraphics[width=0.12\textwidth]{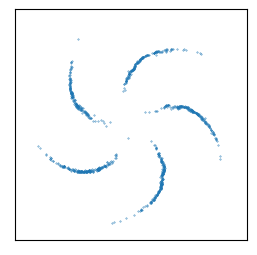}%
}
\subfloat{%
\includegraphics[width=0.12\textwidth]{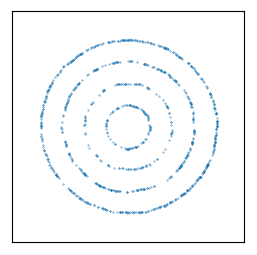}%
}
\subfloat{%
\includegraphics[width=0.12\textwidth]{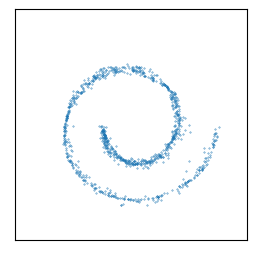}%
}
\caption{Estimated densities and generated samples using Entropy-Kantorovic
potentials, on 2D examples. (Top row) Ground-truth log-densities; (Middle row) Our approximated log-density $\rho_0^\ep$; (Bottom row) Generated samples flowing from standard Normal.}
\label{fig:examples}
\end{figure}

In Figure \ref{fig:examples}, we present the ground-truth log-densities, our
estimated log-densities, and generated samples flowing from a standard Normal
distribution to the target. The added blur in our estimated log-densities
highlights the effect of the entropic interpolation (we trained with $\ep=1$), though the
generated samples seem largely unaffected, and are well-concentrated.

\section{Discussion and future work}
We have presented a novel framework for  density estimation and generative
modelling with CNFs, based on well-establish results from entropy-regularized optimal
transport. Rather than solving a dynamic problem, we exploit a dual formulation
that easily takes advantage of the function-approximation abilities of neural
networks. This allows us to define the estimated densities and their normalizing flows in (near)
closed form. We studied toy problems, but the method we have presented readily
extends to higher-dimensions, which we leave for future work.


\bibliographystyle{plain}
\bibliography{bib}

\cleardoublepage
\setcounter{equation}{0}
\setcounter{figure}{0}
\setcounter{table}{0}
\setcounter{page}{1}

\makeatletter

%

 \hsize\textwidth
    \linewidth\hsize
    \vskip -0.6in
    \@toptitlebar
    \begin{center}
    {\LARGE\bf Suplementary material: Learning normalizing flows from Entropy-Kantorovich
potentials\par}
    \end{center}
    \@bottomtitlebar

\appendix
\addcontentsline{toc}{section}{Appendices}
\renewcommand{\thesubsection}{\Alph{subsection}}

\section{Some concepts from Optimal Transport Theory}\label{app:OT}
 
 We briefly introduce the Monge problem in $\Rd$ for the distance square cost
 function and highlight the relation with the Kantorovich relaxation.
 For more detail, see e.g. \cite{santambrogio2015optimal, vil08}.
 First, let us recall the definition of push-forward of a measure. 
 
 \subsection*{The push-forward of a measure}
 Let $T: \Rd \to\Rd$ be a Borel function and $\mu,\nu$ be probability measures in $\Rd$. The push-forward measure $T_\sharp\mu \in \Pro(\Rd)$ is defined by 
 \begin{equation}\label{eq:PF_set}
    T_\sharp\mu(A) := \mu(T^{-1}(A)) = \mu\left(\left\{x \in \Rd : T(x)\in A \right\}\right) \text{ for any Borel measurable set }A \subset \Rd.
\end{equation}
\
Equivalently, one can write $T_\sharp\mu$ in integral terms
\begin{equation}
    \int_{\Rd} h(y) d T_{\sharp}\mu (y) = \int_{\Rd} h(T(x)) d \mu(x), \quad \forall \, h:\Rd \to \mathbb{R} \text{ Borel}.
\end{equation}
In particular, if we assume $T$ additionally differentiable, $T_{\sharp}\mu = \nu$ can be simply written as classical change of variables formula
\begin{equation}\label{app:eqpushC1}
\mu(x) = \nu(T(x))\vert \mathrm{det}\,\mathrm{J}_T(x)\vert.
\end{equation}
Then, by applying the $\log$ in both sides in $\eqref{app:eqpushC1}$ one has 
$$ \log \mu(x) = \log \nu(T(x)) + \log \vert \mathrm{det}\,\mathrm{J}_T(x)\vert,$$
 where $\mathrm{J}_T$ denotes the Jacobian of a map $T$.
\subsection*{Monge problem and its Kantorovich relaxation}
 The Monge problem seeks to find the best optimal transport map \textit{transporting} $\mu$ and $\nu$, i.e. $T_{\sharp}\mu=\nu$, that minimizes the total work \begin{equation}\label{app:eqMonge}
 \inf_{T_{\sharp}\mu=\nu} \int_{\Rd}\frac{1}{2}\Vert x-T(x)\Vert^2d\mu(x) =  \inf\left\lbrace \int_{\Rd}\frac{1}{2}\Vert x-T(x)\Vert^2d\mu(x) : T:\Rd\to\Rd \mbox{ Borel and } T_{\sharp}\mu=\nu \right\rbrace.
 \end{equation}
 In general, the problem \eqref{app:eqMonge} does not always admit a minimizer. The class of functions $\mathcal{T}(\mu,\nu) = \lbrace T:\Rd\to\Rd : T_{\sharp}\mu=\nu$  and $T$  Borel $\rbrace$ can even be empty. It is enough to take, for example $\mu = \delta_{x_1}$ and $\nu = \frac{1}{2}\delta_{y_1}+\frac{1}{2}\delta_{y_2}$. 
 
The Kantorovich \textit{relaxation} instead 
 \begin{equation}\label{app:eqKanto}
\min_{\gamma\in\Pi(\mu,\nu)}\mathcal{C}(\gamma) := \min_{\gamma\in\Pi(\mu,\nu)}\int_{\Rd\times\Rd}\frac{1}{2}\Vert x-y\Vert^2 d\gamma(x,y),
 \end{equation}
admits a minimizer, since the set $\Pi(\mu,\nu)$ is compact and the cost function $\mathcal{C}(\gamma)$ is lower semi-continuous in the weak$^*$-topology (convergence in law). Notice that the set of transport maps $\mathcal{T}(\mu,\nu)$ can be identified with a subset of $\Pi(\mu,\nu)$ by writing for every $T\in \mathcal{T}(\mu,\nu)$,  $\gamma_T = ({\rm Id}, T)_{\sharp}\mu \in \Pi(\mu,\nu)$. Then,
$$ \min_{\gamma\in\Pi(\mu,\nu)}\int_{\Rd\times\Rd}\frac{1}{2}\Vert x-y\Vert^2 d\gamma(x,y) \leq  \inf_{T_{\sharp}\mu=\nu} \int_{\Rd}\frac{1}{2}\Vert x-T(x)\Vert^2d\mu(x). $$
Under some hypothesis on $\mu$ and $\nu$, one can also show that the equality holds in the above equation. In other words, the solution of \eqref{app:eqKanto} is of \textit{Monge-type}, $\gamma_T = ({\rm Id},T)_{\sharp}\mu$. This is precisely the statement of Brenier's Theorem.
 
 \begin{thm}[Brenier]\label{thm:brenier}
Let $\mu$ and $\nu$ be Borel probability measures on $\R^n$, $c(x,y) = \frac{1}{2}\|x-y\|^2$ be a cost function and suppose $\mu$ has a density with respect to Lebesgue. Then the optimal plan $\gamma$ solving \eqref{app:eqKanto} is supported on the graph of a map $T:\R^d \to \R^n$ satisfying $T_\sharp \mu = \nu$ (i.e. $T\in \mathcal{T}(\mu, \nu)$), i.e. $\gamma = (\operatorname{Id}, T)_\sharp \mu$. Moreover, this map is unique and there exists a convex function $u$ such that $T(x)= \nabla u(x)$.  
\end{thm}
 
 As a consequence, the Monge problem \eqref{app:eqMonge} admits a unique minimizer. 
 
 A natural question is to enquire when the optimal map $T$ in \eqref{app:eqMonge} is differentiable, allowing us to write the condition $T_{\sharp}\mu = \nu$ as in \eqref{app:eqpushC1}. One theoretical and insightful result due to Caffarelli guarantees the regularity of the potentials $u$. Assume that $\mu$ has compact support  and $\nu$ has finite second moments. Then, at least when $\mu(x) = \exp(-W(x)-\vert x\vert^2)\dy$ and $\nu(y) = \exp(V(y)-\vert y\vert^2)\dx$ with $V,W$ convex, the map $T = \nabla u$ is $1$-Lipschitz and $\nu = T_{\sharp}\mu$ \cite{Caf00,Caf02,Kol13}.

\section{Absolutely continuous curves and geodesics in $\mathbb{W}_2$}\label{app:geodesics}
Let $\rho(t)$ be a curve in $\Pro(\Rd)$, i.e. $\rho:[0,1]\to\Pro(\Rd)$, the metric derivative of $\rho(t)$ denoted by $\vert \dot{\rho}\vert(t)$ is defined by $$
 \vert \dot{\rho}\vert(t) = \lim_{h\to0^+}\dfrac{W_2(\rho(t+h),\rho(t))}{h} \quad \mbox{ provided the limit exists}. $$
The following theorems guarantee the existence of the metric derivative for Lipschitz curves in $(\Pro_2(\Rd),W_2)$ and relate absolutely continuous curves in $(\Pro_2(\Rd),W_2)$ with solutions of the continuity equation. 
 We refer to \cite{AmbGigSav} for the prove and further details.
 
\begin{thm} Suppose that $\rho:[0,1]\to\Pro(\Rd)$ is Lipschitz continuous, i.e. for all $s,t\in[0,1]$, $W_2(\rho_s,\rho_t)\leq L\vert t-s\vert,$ for $L>0$. Then the metric derivative $\vert \dot{\rho}\vert(t)$ exists for almost every $t\in[0,1]$. Moreover, for all $t<s$ 
$$
W_2(\rho(t),\rho(s)) \leq \int^s_t \vert \dot{\rho}\vert(a)da.
$$
\end{thm}
 
\begin{defin} A curve $\rho:[0,1]\to\Pro(\Rd)$ is said to be absolutely continuous if there exists a function $f$ such that
$$
W_2(\rho(t),\rho(s)) \leq \int^s_t f(a)da, \quad \forall \  s < t.
$$
\end{defin} 
   
The next theorems relates the continuity equation and an ODE flows constructed in this paper. We refer to \cite{santambrogio2015optimal} for the proofs and in-depth discussion of the results.   
   
\begin{thm} Let $(\rho_t)_{t\in[0,1]}$ be an absolutely continuous curve in $(\Pro_2(\Rd),W_2)$. Then, there exists a vector field $v_t\in L^2(\rho_t,\Rd)$ such that the continuity equation $\partial_t \rho_t + \nabla\cdot (v_t\rho_t) = 0$ is satisfied in the weak sense and, for almost every $t\in[0,1]$, $\vert v_t\vert_{L^2(\rho_t)} \leq \vert \dot{\rho}\vert(t)$. Moreover, the converse also holds: if $(\rho_t)_{t\in[0,1]}$ is a curve in $(\Pro_2(\Rd),W_2)$, $v_t \in L^2(\Rd,\rho_t)$ such that $\int^1_0\int_{\Omega}\vert v_t\vert^2\rho_t\dx \dt < +\infty$ solving $\partial_t\rho_t + \nabla\cdot(v_t\rho_t) = 0$, then $\rho_t$ is absolutely continuous in $W_2$ and for almost every $t\in[0,1], \ \vert \dot{\rho}\vert(t)\leq \vert v_t\vert_{L^2(\rho_t)}$.
\end{thm}   

\begin{defin}
A curve $\rho:[0,1]\to X$ is said to be a geodesic between $\mu$ and $\nu\in X$ if it minimizes the length among all curves such that $\rho(0)=\mu$ and $\rho(1)=\nu$. 
\end{defin}
Let us denote by ${\rm L}(\rho)$ the length of a curve  $\rho:[0,1]\to X$, 
$${\rm L}(\rho):=\sup\left\{\sum_{k=0}^{n-1}d(\rho(t_k),\rho(t_{k+1}))\,:\,n\geq 1,\, 0=t_0<t_1<\dots<t_n=1\right\}.$$

A space $(X,d)$ is said to be a {\it geodesic space} if it holds
$$d(\mu,\nu)=\min\{{\rm L}(\rho)\,:\,\rho \mbox{ is absolutely continuous},\,\rho(0)=\mu,\rho(1)=\nu\},$$
i.e. there exist geodesics between arbitrary points.

\begin{prop}[$(\Pro_p(\Omega),W_2)$ is a geodesic space]\label{app:propgeo}
Let $\Omega\subset\Rd$ be convex, $\mu,\nu\in\Pro_p(\Omega)$ and $\gamma\in\Pi(\mu,\nu)$ an optimal transport plan for the cost $c(x,y) = \vert x-y\vert^p$, $p\geq 1$. Define the curve $\pi_t:\Omega\times\Omega\to\Omega$ through $\pi_t(x,y) = (1-t)x+ty$. Then the curve $\rho_t = (\pi_t)_{\sharp}\gamma$ is a constant speed geodesic in $(\Pro_p(\Omega),W_p)$ from $\mu$ to $\nu$. In particular, when an optimal transport plan $\gamma=\gamma_T$ is concentrated in a map $T$, the curve $\rho_t = ((1-t){\rm Id}+tT)_{\sharp}\mu$. 
\end{prop}

\begin{prop}
Let $\mu,\nu$ be two densities in $\Pro_p(\Omega), p\geq2$, $\rho_t=(\pi_t)_\#\gamma$ be the geodesic connecting $\mu$ to $\nu$ introduced in Proposition \ref{app:propgeo} and $T_t(x) = (1-t)x + tT(x)$, where $T$ is the optimal transport map between $\mu$ to $\nu$. Then the velocity field $v_t(y)=(T-{\rm Id})(T_t^{-1}(y))$ is well defined on $\spt(\rho_t)$ for each $t\in ]0,1[$ and satisfies
$$\partial_t\rho_t+\nabla\cdot (\rho_tv_t)=0,\quad \Vert v_t\Vert_{L^p(\rho_t)}=\vert \dot{\rho}\vert(t)=W_p(\mu,\nu).$$
\end{prop}

\section{Dynamical formulation of the Entropy-regularized Optimal Transport}\label{app:dynOTep}

The variational problem \eqref{eq:Wep} can be alternatively writen in the \textit{dynamic} formulation \cite{gentil2017,GigTamBB18,LeoSurvey}
\begin{align}\label{app:eqwepdynamical}
W_{\ep}^{2}(&\pd,\pn) = \min_{(\rhoept,w^{\ep}_t)}\int_0^1\int_{\Rd}\dfrac{\Vert w^{\ep}_t\Vert^2}{2}d\rhoept dt + \frac{\ep}{2}\left({\rm H}(\pd) + {\rm H}(\pn)\right),\\
&=\sup_{(\phiept,\psiept)}\int_{\Rd}(\phiep_1-\psiep_1) d\pd + \int_{\Rd}(\phiep_0-\psiep_0) d\pn + \frac{\ep}{2}\left({\rm H}(\pd) + {\rm H}(\pn)\right),
\end{align}
where the minimum must be understood as taken among all couples $(\rhoept,w^{\ep}_t)$ solving the backward and forward Fokker-Planck equations 
\[
-\partial_t \rhoept + \nabla\cdot (\nabla\phiept \rhoept) = \frac{\ep}{2}\Delta \rhoept, \quad \text{ and } \quad \partial_t \rhoept + \nabla\cdot (\nabla\psiept \rhoept) = \frac{\ep}{2}\Delta \rhoept, 
\]
for $t\in[0,1]$ such that $\rho^\epsilon_0=\pd, \rho^\epsilon_1 = \pn$; while the supremum is taking over the couple  $(\phiept,\psiept)$ solving the Hamilton-Jacobi-Bellman equations
\[
\partial_t \phiept = \dfrac{\Vert \nabla \phiept\Vert^2}{2} + \dfrac{\ep}{2}\Delta \phiept, \quad \text{and} \quad -\partial_t \psiept = \dfrac{\Vert \nabla \psiept\Vert^2}{2} + \dfrac{\ep}{2}\Delta \psiept.
\]
The optimal vector field $w^{\ep}_t$ is given by the Entropy-Kantorovich potentials $w^{\ep}_t = \nabla (\phiept-\psiept)/2$, which corresponds to the regularized constant speed geodesic in the $2$-Wasserstein space. 

By writing $w^{\ep}_t = \vept -\ep\nabla\log(\rhoep_t)$, the variational problem \eqref{app:eqwepdynamical} corresponds to eq \eqref{eq: intro_p}
\begin{align}
\frac{\ep}{2}\left({\rm H}(\pd) + {\rm H}(\pn)\right) + \underset{(\rho_t, \vept)}{\inf} \int_0^1\int_{\Rd} \pran{\frac{\|\vept\|^2}{2} + \frac{\ep^2}{8}\| \nabla \log \rho_t \|^2}\rho_t \ dx\ dt,
\end{align}
where $(\rho^{\ep}_t, \vept)$ is such that $\rhoep_{0} = \pd$, $\rho_1 = \pn$ and solves the continuity equation 

In the following, we give a formal computation explaining the optimal conditions obtained via the above primal-dual relation.

\subsection*{Characterization \eqref{app:eqwepdynamical} via primal-dual problems} 

Let us assume that $\varphi_t$ and $\psi_t$ solves the respective HJB equations
and define $\alpha_t = (\varphi_t-\psi_t)/2$. Let us compute
\begin{align}\label{eq:weakVt}
\dfrac{d}{ds}\bigg|_{s=t}\int_{\R^d}\mu^{\ep}_s \alpha_sdx = \int_{\R^d}\dfrac{d}{ds}\bigg|_{s=t}\mu^{\ep}_s \alpha_sdx +\int_{\R^d}\mu^{\ep}_s\dfrac{d}{ds}\bigg|_{s=t} \alpha_sdx =: {\rm (I)} + {\rm (II)}.
\end{align}

Since $\varphi_t$ and $\psi_t$ solves the respective HJB equations, we have
\begin{align*}
{\rm (II)} = \int_{\R^d}\mu^{\ep}_s\dfrac{d}{ds}\bigg|_{s=t}\alpha_sdx &=
\int_{\R^d}\left(-\dfrac{\Vert \nabla \psi_t\Vert^2}{4}-\dfrac{\Vert
\nabla \varphi_t\Vert^2}{4}-\dfrac{\ep}{4}\Delta(\psi_t+\varphi_t)\right)\mu^{\ep}_s dx \\
&= \int_{\R^d}\left(-\dfrac{\Vert \nabla\psi_t\Vert^2}{4}-\dfrac{\Vert
\nabla \varphi_t\Vert^2}{4}+\dfrac{1}{4}\langle \nabla(\psi_t+\varphi_t),\ep\nabla\log(\mu^{\ep}_s)\rangle \right)\mu^{\ep}_s dx \\ 
&\leq \int_{\R^d}\left(-\dfrac{\Vert \nabla \psi_t\Vert^2}{4}-\dfrac{\Vert
\nabla\varphi_t\Vert^2}{4}+\dfrac{1}{8}\Vert \nabla(\psi_t+\varphi_t)\Vert^2 +\dfrac{\ep^2}{8}\Vert \nabla\log(\mu^{\ep}_s)\Vert^2 \right)\mu^{\ep}_s dx 
\end{align*}
The second line follows from the first by applying integration by parts to pass a gradient onto
the measure $\mu^\ep$, then multiplying and dividing by $\mu^\ep$, and then
using $(\nabla \mu^\ep)/\mu^\ep = \nabla \log \mu^\ep$.
The last line is with equality if and only if $\ep\nabla\log(\mu^{\ep}_t) =
\nabla(\psi_t+\varphi_t)$ almost everywhere. 

Now, if $(\mu^{\ep}_t, v_t)$ solves the continuity equation then
\begin{align}
{\rm (I)} = \int_{\R^d}\dfrac{d}{ds}\bigg|_{s=t}\mu^{\ep}_s\alpha_sdx &=
-\int_{R^d}\alpha_s\nabla\cdot(v_t\mu^{\ep}_t) dx = \int_{\R^d}\langle
\nabla(\psi_t-\varphi_t)/2,v_t\rangle \mu^{\ep}_t dx \\
&\leq \int_{\R^d}\dfrac{1}{2}\Vert \nabla(\psi_t-\varphi_t)/2\Vert^2 + \dfrac{1}{2}\Vert v_t\Vert^2 \mu^{\ep}_t dx,
\end{align}
with equality if and only if $v_t = \nabla(\psi_t-\varphi_t)/2.$ Finally, integrating \eqref{eq:weakVt} over time one has
\begin{align}
\frac{1}{2}\int_{\R^d}(\psi_1-\varphi_1)d\rho_1 +
\frac{1}{2}\int_{\R^d}(\psi_0-\varphi_0)d\rho_0 \leq  \int^1_{0}\int_{\R^d}\frac{\Vert v_t\Vert^2}{2} + \dfrac{\ep}{8}\Vert \nabla \log\mu^{\ep}_t\Vert^2dxdt.
\end{align}

Since all the computations above are arbitrary we have that
\[
\sup_{(\psi_t,\varphi_t)} \left\lbrace
\frac{1}{2}\int_{\R^d}(\psi_1-\varphi_1)d\rho_1 +
\frac{1}{2}\int_{\R^d}(\psi_0-\varphi_0)d\rho_0 : 
\begin{aligned}
      \partial_t \varphi_t = \frac{\Vert \nabla \varphi_t\Vert^2}{2} +
      \frac{\ep}{2}\Delta \varphi_t\\
      -\partial_t \psi_t = \frac{\Vert \nabla \psi_t\Vert^2}{2} + \frac{\ep}{2}\Delta \psi_t 
\end{aligned} \right\rbrace \leq \]
\[
\leq \inf_{(\mu^{\ep}_t,v_t)}\left\lbrace \int^1_{0}\int_{\R^d}\frac{\Vert v_t\Vert^2}{2} + \dfrac{\ep}{8}\Vert \nabla \log\mu^{\ep}_t\Vert^2dxdt \, : \,  \begin{aligned}
      \partial_t \mu^\ep_t + \nabla\cdot (v_t \mu^\ep_t) = 0\\
      \mu^{\ep}_0 = \pd,\mu^{\ep^2}_1 = \pn
\end{aligned} \right\rbrace
\]

The equality is reached when $v_t = \nabla (\psi_t-\varphi_t)/2$ and $\mu^{\ep}_t$ is the entropic interpolation $$\mu^\ep_t := \HH_{t\ep}(e^{\varphi^{\ep}}) \,\HH_{(1-t)\ep}(e^{\psi^{\ep}}).$$

In particular, at the optimal $(\mu^{\ep}_t,v_t)$ we have that
\[
\frac{1}{2}\int_{\R^d}(\psi_1-\varphi_1)\pn\dy +
\frac{1}{2}\int_{\R^d}(\psi_0-\varphi_0)\pd\dx = \int^1_{0}\int_{\R^d}\frac{\Vert v_t\Vert^2}{2} + \dfrac{\ep^2}{8}\Vert \nabla \log\mu^{\ep}_t\Vert^2 \mu^{\ep}_t\dx\dt.
\]

\subsection*{Closed-form solutions for $d$-dimensional Gaussians}

\begin{figure}[H]
\captionsetup[subfigure]{labelformat=empty}
\centering
\subfloat{%
\includegraphics[width=0.15\textwidth]{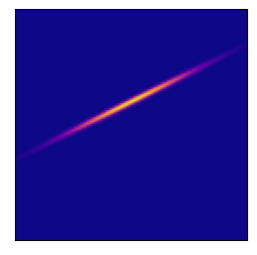}%
}
\subfloat{%
\includegraphics[width=0.15\textwidth]{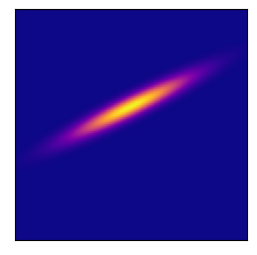}%
}
\subfloat{%
\includegraphics[width=0.15\textwidth]{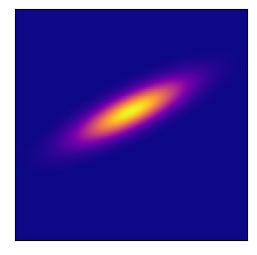}%
}
\subfloat{%
\includegraphics[width=0.15\textwidth]{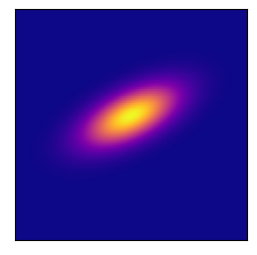}%
}
\subfloat{%
\includegraphics[width=0.15\textwidth]{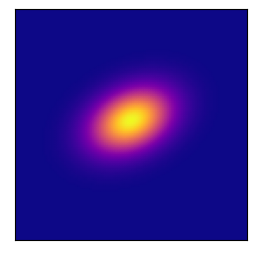}%
}
\subfloat{%
\includegraphics[width=0.15\textwidth]{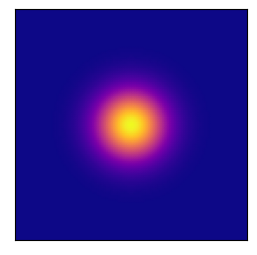}%
}\vspace{-4mm}\\
\subfloat[$t=0$]{%
\includegraphics[width=0.15\textwidth]{figures/gaussian_flow/pt0.png}%
}
\subfloat[$t=0.2$]{%
\includegraphics[width=0.15\textwidth]{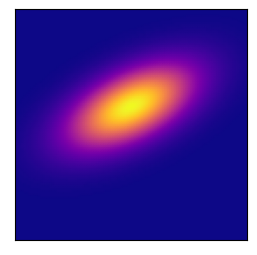}%
}
\subfloat[$t=0.4$]{%
\includegraphics[width=0.15\textwidth]{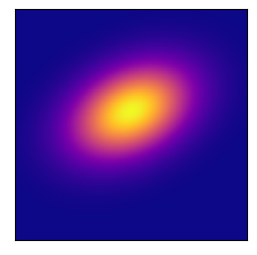}%
}
\subfloat[$t=0.6$]{%
\includegraphics[width=0.15\textwidth]{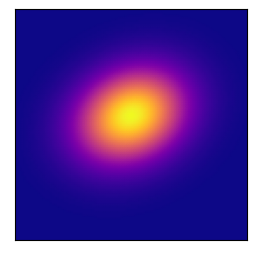}%
}
\subfloat[$t=0.8$]{%
\includegraphics[width=0.15\textwidth]{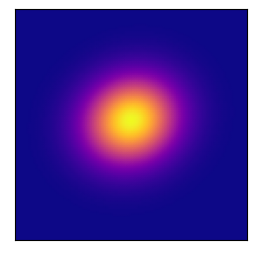}%
}
\subfloat[$t=1$]{%
\includegraphics[width=0.15\textwidth]{figures/gaussian_flow/pt1.png}%
}

\caption{Flow between two Normal distributions: the source distribution has a
  degenerate covariance structure and the target is the standard Normal distribution. (Top) $W_2$ geodesics
(Bottom) The entropy-regularized interpolation.}
\label{fig:gaussians}
\end{figure}

We illustrate in the following example and accompanied Figure
\ref{fig:gaussians} the smoothing effect of the regularization on
$2$-Wasserstein geodesics for two Gaussian distributions. In the example, we
notice that the initial distribution has a degenerate covariance structure that
is maintained in the $2$-Wasserstein case. When incorporating regularization, we
see a smoothed out distribution that more closely resembles the target throughout the flow.

\begin{exo}[Comparing geodesics and entropic interpolation for Gaussian
  distributions]\label{ex:gaussians}
Consider two two multivariate Gaussian distributions $\rho_0 = \mathcal{N}(m_0,\Sigma_0)$ and $\rho_1 = \mathcal{N}(m_1,\Sigma_1)$. The geodesics under the Wasserstein metric is given by $\rho_t = \mathcal{N}(m_t, \Sigma_t)$ {\rm \cite{McC97}} with $m_t = (1-t)m_0 + tm_1$ and $$\Sigma_t = (1-t)^2\Sigma_0 + t^2\Sigma_1 + t(1-t)[(\Sigma_0\Sigma_1)^{1/2} + (\Sigma_1\Sigma_0)^{1/2}].$$
The entropic interpolation is $\rhoep_t = \left(m_t, \Sigma^{\ep}_t\right)$ 
$t\in[0,1]$, where $m_t$ is the same as before, and 
$$
\Sigma^{\ep}_t =  (1-t)^2\Sigma_0 + t^2\Sigma_1 + t(1-t)
    \left[\left(\frac{\ep^2}{16}I + \Sigma_0\Sigma_1\right)^{1/2}\right.+\left. \left(\frac{\ep^2}{16}I + \Sigma_1\Sigma_0\right)^{1/2}\right].
$$
Notice that the covariance structures are the same up to a function of $\varepsilon$ that appears in the mixing term. Clearly $\Sigma^{\ep}_t\to\Sigma_t$ when $\ep\to 0$.
\end{exo}

\section{Algorithm details and hyperparameters}\label{sec: algo_main}
For all the experiments, we use four fully connected linear layers with ReLU activations. The hidden dimension of the layers was 64. 

The 2D datasets considered are: `Checkerboard', `Swissroll', `Rings' (four concentric rings), `Moons', `Circles' (two concentric rings), `2spirals', `Pinwheel', and `8gaussians'.

For \textit{training}, the following hyperparameters are constant across all datasets:
\begin{itemize}
\item Batch-size for the sampled data (and sampling from the Normal distribution) was 1000
\item Number of samples for the Monte-Carlo (MC) integration was 100
\item Learning rate for stochastic gradient descent was $10^{-3}$
\item $L_2$ penalty term to enforce the $(c,\varepsilon)$-transformation was $10^{-5}$
\end{itemize} 
For all but the `Rings' dataset, the number of iterations was 20000 --- for `Rings', we needed to use 40000 iterations.

Finally, for \textit{generating} samples, we used a batch-size of 1000, 200 MC
samples, and used the default Dormand-Prince Runge-Kutta 4(5) adaptive solver (\texttt{dopri5}) ODE integrator from the \texttt{torchdiffeq} Python package \cite{Che18}.

\end{document}